\documentclass{article}

\PassOptionsToPackage{numbers, compress}{natbib}
\usepackage[preprint]{neurips_style}

\usepackage[utf8]{inputenc} 
\usepackage[T1]{fontenc}    
\usepackage{hyperref}       
\usepackage{url}            
\usepackage{booktabs}       
\usepackage{amsfonts}       
\usepackage{nicefrac}       
\usepackage{microtype}      
\usepackage{xcolor}         

\usepackage{amssymb}
\usepackage{statmath}

\usepackage{amsmath,amssymb,amsthm}

\usepackage{pifont}
\newcommand{\xmark}{\ding{55}}

\renewcommand{\gg}{\mathbf{g}}

\newenvironment{talign*}
{\csname align*\endcsname}
{\endalign}

\usepackage[utf8]{inputenc}         %
\usepackage[T1]{fontenc}            %
\usepackage{url}                    %
\usepackage{booktabs}               %
\usepackage{amsfonts}               %
\usepackage{nicefrac}               %
\usepackage{microtype}              %
\usepackage{xcolor}                 %
\usepackage{algpseudocode}
\usepackage{algorithm}
\usepackage{graphicx}
\usepackage{subcaption}
\usepackage[flushleft]{threeparttable}
\usepackage{float}
\usepackage{multirow}
\usepackage{xspace}
\usepackage{natbib}
\usepackage{enumitem}
\usepackage[font=small]{caption}
\usepackage{autobreak}
\usepackage{sidecap}
\usepackage{wrapfig}
\usepackage{bbding}
\usepackage[toc, page, header]{appendix}
\usepackage{tikz}
\usepackage{xcolor}
\usepackage{pifont}
\usepackage{mdframed}
\usepackage{colortbl}
\usepackage[english]{babel}
\usepackage{hyperref}

\hypersetup{
  colorlinks=true,
  linkcolor=blue,
  citecolor=blue,
  urlcolor=blue
}

\definecolor{coral}{RGB}{255,127,80}
\definecolor{darkgreen}{RGB}{0,100,0}
\definecolor{darkyellow}{RGB}{204,153,0}
\definecolor{salmon}{RGB}{250,128,114}
\definecolor{darkred}{RGB}{150,0,0}

\newcommand{\transparentred}[1]{%
  \tikz[baseline=(X.base)] \node[fill=red, fill opacity=0.1, text opacity=1, inner sep=2pt, outer sep=0pt] (X) {#1};%
}
\newcommand{\thmref}[1]{\hyperref[#1]{\transparentred{Theorem~\ref*{#1}}}}

\newcommand{\transparentgray}[1]{%
  \tikz[baseline=(X.base)] \node[fill=gray, fill opacity=0.1, text opacity=1, inner sep=2pt, outer sep=0pt] (X) {#1};%
}
\newcommand{\defref}[1]{\hyperref[#1]{\transparentgray{Definition~\ref*{#1}}}}

\newcommand{\transparentblue}[1]{%
  \tikz[baseline=(X.base)] \node[fill=blue, fill opacity=0.1, text opacity=1, inner sep=2pt, outer sep=0pt] (X) {#1};%
}
\newcommand{\propref}[1]{\hyperref[#1]{\transparentblue{Proposition~\ref*{#1}}}}

\newcommand{\transparentgreen}[1]{%
  \tikz[baseline=(X.base)] \node[fill=green, fill opacity=0.1, text opacity=1, inner sep=2pt, outer sep=0pt] (X) {#1};%
}
\newcommand{\assumpref}[1]{\hyperref[#1]{\transparentgreen{Assumption~\ref*{#1}}}}

\newcommand{\transparentyellow}[1]{%
  \tikz[baseline=(X.base)] \node[fill=yellow, fill opacity=0.1, text opacity=1, inner sep=2pt, outer sep=0pt] (X) {#1};%
}
\newcommand{\remarkref}[1]{\hyperref[#1]{\transparentyellow{Remark~\ref*{#1}}}}

\newcommand{\transparentpurple}[1]{%
  \tikz[baseline=(X.base)] \node[fill=purple, fill opacity=0.1, text opacity=1, inner sep=2pt, outer sep=0pt] (X) {#1};%
}
\newcommand{\hypref}[1]{\hyperref[#1]{\transparentpurple{Hypothesis~\ref*{#1}}}}

\newcommand{\transparentorange}[1]{%
  \tikz[baseline=(X.base)] \node[fill=orange, fill opacity=0.1, text opacity=1, inner sep=2pt, outer sep=0pt] (X) {#1};%
}
\newcommand{\conjref}[1]{\hyperref[#1]{\transparentorange{Conjecture~\ref*{#1}}}}

\newcommand{\transparentcyan}[1]{%
  \tikz[baseline=(X.base)] \node[fill=cyan, fill opacity=0.1, text opacity=1, inner sep=2pt, outer sep=0pt] (X) {#1};%
}
\newcommand{\lemref}[1]{\hyperref[#1]{\transparentcyan{Lemma~\ref*{#1}}}}

\newcommand{\transparentmagenta}[1]{%
  \tikz[baseline=(X.base)] \node[fill=magenta, fill opacity=0.1, text opacity=1, inner sep=2pt, outer sep=0pt] (X) {#1};%
}
\newcommand{\corref}[1]{\hyperref[#1]{\transparentmagenta{Corollary~\ref*{#1}}}}

\newcommand{\transparentlime}[1]{%
  \tikz[baseline=(X.base)] \node[fill=lime, fill opacity=0.1, text opacity=1, inner sep=2pt, outer sep=0pt] (X) {#1};%
}
\newcommand{\exampleref}[1]{\hyperref[#1]{\transparentlime{Example~\ref*{#1}}}}

\newcommand{\transparentpink}[1]{%
  \tikz[baseline=(X.base)] \node[fill=pink, fill opacity=0.1, text opacity=1, inner sep=2pt, outer sep=0pt] (X) {#1};%
}
\newcommand{\noteref}[1]{\hyperref[#1]{\transparentpink{Notation~\ref*{#1}}}}

\newcommand{\transparentviolet}[1]{%
  \tikz[baseline=(X.base)] \node[fill=violet, fill opacity=0.1, text opacity=1, inner sep=2pt, outer sep=0pt] (X) {#1};%
}
\newcommand{\claimref}[1]{\hyperref[#1]{\transparentviolet{Claim~\ref*{#1}}}}

\newcommand{\transparentsalmon}[1]{%
  \tikz[baseline=(X.base)] \node[fill=salmon, fill opacity=0.1, text opacity=1, inner sep=2pt, outer sep=0pt] (X) {#1};%
}
\newcommand{\probref}[1]{\hyperref[#1]{\transparentsalmon{Problem~\ref*{#1}}}}

\newcommand{\transparentlavender}[1]{%
  \tikz[baseline=(X.base)] \node[fill=lavender, fill opacity=0.1, text opacity=1, inner sep=2pt, outer sep=0pt] (X) {#1};%
}
\newcommand{\obsref}[1]{\hyperref[#1]{\transparentlavender{Observation~\ref*{#1}}}}

\newcommand{\transparentteal}[1]{%
  \tikz[baseline=(X.base)] \node[fill=teal, fill opacity=0.1, text opacity=1, inner sep=2pt, outer sep=0pt] (X) {#1};%
}
\newcommand{\figref}[1]{\hyperref[#1]{\transparentteal{Figure~\ref*{#1}}}}

\newcommand{\transparentdarkgreen}[1]{%
  \tikz[baseline=(X.base)] \node[fill=darkgreen, fill opacity=0.1, text opacity=1, inner sep=2pt, outer sep=0pt] (X) {#1};%
}
\newcommand{\tabref}[1]{\hyperref[#1]{\transparentdarkgreen{Table~\ref*{#1}}}}

\newcommand{\transparentdarkyellow}[1]{%
  \tikz[baseline=(X.base)] \node[fill=darkyellow, fill opacity=0.1, text opacity=1, inner sep=2pt, outer sep=0pt] (X) {#1};%
}
\newcommand{\secref}[1]{\hyperref[#1]{\transparentdarkyellow{Section~\ref*{#1}}}}

\newcommand{\transparentcoral}[1]{%
  \tikz[baseline=(X.base)] \node[fill=coral, fill opacity=0.1, text opacity=1, inner sep=2pt, outer sep=0pt] (X) {#1};%
}
\newcommand{\appref}[1]{\hyperref[#1]{\transparentcoral{Appendix~\ref*{#1}}}}
\newcommand{\algoref}[1]{\hyperref[#1]{\transparentteal{Algorithm~\ref*{#1}}}}

\newtheoremstyle{custom}
{1pt} %
{1pt} %
{\itshape} %
{} %
{\bfseries} %
{} %
{ } %
{\thmname{#1} \thmnumber{#2} \thmnote{(#3)} . } %

\theoremstyle{custom}

\newtheorem{innerhypothesis}{Hypothesis}
\newtheorem{innerconjecture}{Conjecture}

\newtheorem{innerexample}{Example}
\newtheorem{innernotation}{Notation}
\newtheorem{innerclaim}{Claim}
\newtheorem{innerproblem}{Problem}

\newtheorem{innerobservation}{Observation}

\newmdenv[
  backgroundcolor=gray!10,
  linecolor=gray!100,
  linewidth=0.8pt,
  skipabove=2pt,
  skipbelow=2pt,
  innertopmargin=10pt,
  innerbottommargin=5pt,
  innerleftmargin=5pt,
  innerrightmargin=5pt,
]{definitionframe}

\newmdenv[
  backgroundcolor=blue!10,
  linecolor=blue!100,
  linewidth=0.8pt,
  skipabove=2pt,
  skipbelow=2pt,
  innertopmargin=10pt,
  innerbottommargin=5pt,
  innerleftmargin=5pt,
  innerrightmargin=5pt,
]{propositionframe}

\newmdenv[
  backgroundcolor=green!10,
  linecolor=green!100,
  linewidth=0.8pt,
  skipabove=2pt,
  skipbelow=2pt,
  innertopmargin=10pt,
  innerbottommargin=5pt,
  innerleftmargin=5pt,
  innerrightmargin=5pt,
]{assumptionframe}

\newmdenv[
  backgroundcolor=yellow!10,
  linecolor=yellow!100,
  linewidth=0.8pt,
  skipabove=2pt,
  skipbelow=2pt,
  innertopmargin=10pt,
  innerbottommargin=5pt,
  innerleftmargin=5pt,
  innerrightmargin=5pt,
]{remarkframe}

\newmdenv[
  backgroundcolor=red!10,
  linecolor=red!100,
  linewidth=0.8pt,
  skipabove=2pt,
  skipbelow=2pt,
  innertopmargin=10pt,
  innerbottommargin=5pt,
  innerleftmargin=5pt,
  innerrightmargin=5pt,
]{theoremframe}

\newmdenv[
  backgroundcolor=purple!10,
  linecolor=purple!100,
  linewidth=0.8pt,
  skipabove=2pt,
  skipbelow=2pt,
  innertopmargin=10pt,
  innerbottommargin=5pt,
  innerleftmargin=5pt,
  innerrightmargin=5pt,
]{hypothesisframe}

\newmdenv[
  backgroundcolor=orange!10,
  linecolor=orange!100,
  linewidth=0.8pt,
  skipabove=2pt,
  skipbelow=2pt,
  innertopmargin=10pt,
  innerbottommargin=5pt,
  innerleftmargin=5pt,
  innerrightmargin=5pt,
]{conjectureframe}

\newmdenv[
  backgroundcolor=cyan!10,
  linecolor=cyan!100,
  linewidth=0.8pt,
  skipabove=2pt,
  skipbelow=2pt,
  innertopmargin=10pt,
  innerbottommargin=5pt,
  innerleftmargin=5pt,
  innerrightmargin=5pt,
]{lemmaframe}

\newmdenv[
  backgroundcolor=magenta!10,
  linecolor=magenta!100,
  linewidth=0.8pt,
  skipabove=2pt,
  skipbelow=2pt,
  innertopmargin=10pt,
  innerbottommargin=5pt,
  innerleftmargin=5pt,
  innerrightmargin=5pt,
]{corollaryframe}

\newmdenv[
  backgroundcolor=lime!10,
  linecolor=lime!100,
  linewidth=0.8pt,
  skipabove=2pt,
  skipbelow=2pt,
  innertopmargin=10pt,
  innerbottommargin=5pt,
  innerleftmargin=5pt,
  innerrightmargin=5pt,
]{exampleframe}

\newmdenv[
  backgroundcolor=pink!10,
  linecolor=pink!100,
  linewidth=0.8pt,
  skipabove=2pt,
  skipbelow=2pt,
  innertopmargin=10pt,
  innerbottommargin=5pt,
  innerleftmargin=5pt,
  innerrightmargin=5pt,
]{notationframe}

\newmdenv[
  backgroundcolor=violet!10,
  linecolor=violet!100,
  linewidth=0.8pt,
  skipabove=2pt,
  skipbelow=2pt,
  innertopmargin=10pt,
  innerbottommargin=5pt,
  innerleftmargin=5pt,
  innerrightmargin=5pt,
]{claimframe}

\newmdenv[
  backgroundcolor=salmon!10,
  linecolor=salmon!100,
  linewidth=0.8pt,
  skipabove=2pt,
  skipbelow=2pt,
  innertopmargin=10pt,
  innerbottommargin=5pt,
  innerleftmargin=5pt,
  innerrightmargin=5pt,
]{problemframe}

\newmdenv[
  backgroundcolor=lavender!10,
  linecolor=lavender!100,
  linewidth=0.8pt,
  skipabove=2pt,
  skipbelow=2pt,
  innertopmargin=10pt,
  innerbottommargin=5pt,
  innerleftmargin=5pt,
  innerrightmargin=5pt,
]{observationframe}

\usepackage[textwidth=3.5cm]{todonotes}

\newcommand{\ours}{\texttt{PIXAR-DG}\xspace}

\renewcommand{\gg}{>\!\!\!>}

\definecolor{red1}{RGB}{200,30,30}

\newcommand{\hl}[1]{\cellcolor[HTML]{D5E8D4}#1}
\newcommand{\hlB}[1]{\cellcolor[HTML]{F8CECC}#1} 
\newcommand{\hltext}[1]{\colorbox[HTML]{D5E8D4}{#1}}
\newcommand{\hlBtext}[1]{\colorbox[HTML]{F8CECC}{#1}}
\title{Simple Domain Generalization for Strong Pixel-Level Image Tampering Detection in Modern VLMs}

\author{%
  \bfseries Yi Tang$^{1,\star}$\quad Xinyi Shang$^{1,2,\star}$\quad Jiacheng Cui$^{1}$\quad Sondos Mahmoud Bsharat$^{1}$\\[3pt]
  \bfseries Jiacheng Liu$^{1}$ \quad  Xiaohan Zhao$^{1}$\quad Tran Dinh Tien$^{1}$ \quad Ahmed Elhagry$^{1}$ \\[3pt]
  \bfseries  Salwa K. Al Khatib$^{1}$ \quad Tianjun Yao$^{1}$\quad Yonina C. Eldar$^{3}$ \quad Jing-Hao Xue$^{2}$\\[3pt]
  \bfseries Hao Li$^{1}$\quad Salman Khan$^{1}$\quad Zhiqiang Shen$^{1,\dagger}$\\[6pt]
  $^{1}$Mohamed bin Zayed University of Artificial Intelligence\\[1pt]
  $^{2}$University College London\quad $^{3}$Weizmann Institute of Science\\[4pt]
  $^{\star}$Equal Contribution\qquad $^{\dagger}$Corresponding author%
}

\begin{document}

\maketitle

\begin{abstract}

Modern vision-language models (VLMs) have significantly improved image generation and editing capabilities, making pixel-level image tampering detection increasingly important yet challenging under cross-model and out-of-distribution shifts. This work studies domain generalization for pixel-level image tampering detection in modern VLMs like ChatGPT, Gemini, Qwen-Image, etc., aiming to learn tampering localization models that remain robust across diverse VLM-generated manipulation distributions. We propose a simple yet effective domain-generalized training framework built on two practical strategies. First, we introduce a balanced minibatch sampling scheme that strategically samples tampered and real images in each minibatch, preventing biased optimization toward either manipulated artifacts or clean-image priors and avoiding training collapse\footnote{We frequently observe this behavior when the proposed sampling strategy is not used, and we analyze it in our experiments.}, ensuring that each optimization step receives proper sampled gradient signals. Second, we adopt a simple late-injection strategy, where the detector is first trained on large-scale base data until stable convergence, and then exposed to a small amount of newly selected supporting data from emerging VLM distributions, improving adaptability without overfitting to limited new domains. 
Together, these components provide a simple yet strong recipe for improving pixel-level tampering localization and OOD robustness across modern VLMs. Despite the conceptual simplicity, our framework outperforms the prior state-of-the-art PIXAR by a large margin of $26.1\%$ and $26.8\%$ relative improvement in average gIoU and cIoU, respectively, across OOD VLMs of GPT-Images-2.0, Gemini-3.1, FLUX.2, and Seedream 4.5.
Our code is available at \url{https://github.com/VILA-Lab/PIXAR-DG}.

\end{abstract}

\section{Introduction}

Modern vision-language models (VLMs)~\cite{wu2025qwen,gemini25,xia2024gsva} have rapidly advanced image generation and editing, enabling users to manipulate localized visual content with increasingly realistic results. While these capabilities are useful for creative applications, they also raise serious concerns for visual misinformation, content authenticity, and forensic verification. In this context, pixel-level image tampering detection~\cite{shang2026masks} is especially important because it not only predicts whether an image has been manipulated, but also localizes the exact tampered regions.

However, robust pixel-level tampering detection remains challenging under real-world deployment conditions. Modern VLMs differ substantially in architecture, training data, editing interface, generation pipeline, and post-processing behavior, causing large domain shifts across manipulated images from different models. A detector trained on one or several source VLMs may therefore overfit to model-specific artifacts and fail on unseen VLMs, especially when the target distribution comes from a newly released or closed-source model.

This work studies domain generalization for pixel-level image tampering detection in modern VLMs, introducing \ours, a completely new task and problem in the community. Our goal is to learn tampering localization models that capture transferable domain-robust cues rather than superficial artifacts tied to a specific generator or editing domain. To this end, we focus on practical training strategies that improve cross-domain robustness while remaining simple to implement and compatible with existing segmentation-based tampering detectors. We discover that careful source-domain composition and optimization scheduling are critical for VLM tampering detectors, and derive a simple but effective recipe for generator-agnostic localization.

First, we propose balanced minibatch training with both tampered and real images. Each minibatch is constructed to contain an optimal mixture of manipulated and clean samples, preventing the model from being biased toward either tampering artifacts or real-image priors. When combined with multi-domain source data, this strategy encourages every optimization step to see diverse real/tampered contrasts across different VLM distributions, improving cross-domain robustness.
Second, we introduce a late-injection training strategy for incorporating small newly selected distributions. The detector is first trained on a large-scale base dataset until sufficiently stable and generalizable representations are learned. Only after this stage is done, we inject a small amount of new-domain data together with the previous domain data, allowing the detector to adapt to emerging VLM distributions without letting limited new data dominate early feature learning or cause severe overfitting.
This is particularly useful for modern VLM tampering detection, where new model outputs may be available only in small quantities. Our overall training framework provides a simple yet strong recipe for improving pixel-level tampering localization and OOD robustness, beating all prior methods.

Our contributions of this work are: 

\begin{itemize}
\item  We present a practical training framework for domain-generalized pixel-level image tampering detection. Unlike existing methods that often overfit to domain-specific artifacts from particular generators or training distributions, our framework is designed to learn transferable tampering-related cues that remain effective under cross-model and out-of-distribution shifts. The proposed framework is simple to implement, achieving superior performance.
\item We propose a simple balanced real/tampered minibatch training strategy, a late-injection mechanism for small emerging distributions, and a low-learning-rate adaptation schedule that improves pixel-level localization and OOD robustness across modern VLMs.
\item Our method achieves stronger results across various models and domains with substantially less training data. On the PIXAR dataset, our method uses only $19.2\%$ of the original dataset scale, yet consistently improves performance across all new domains, with relative gains of $26.1\%$ and $26.8\%$ in gIoU and cIoU compared with prior state-of-the-art methods. 
\end{itemize}

\begin{figure}[t]
    \centering
    \includegraphics[width=\linewidth]{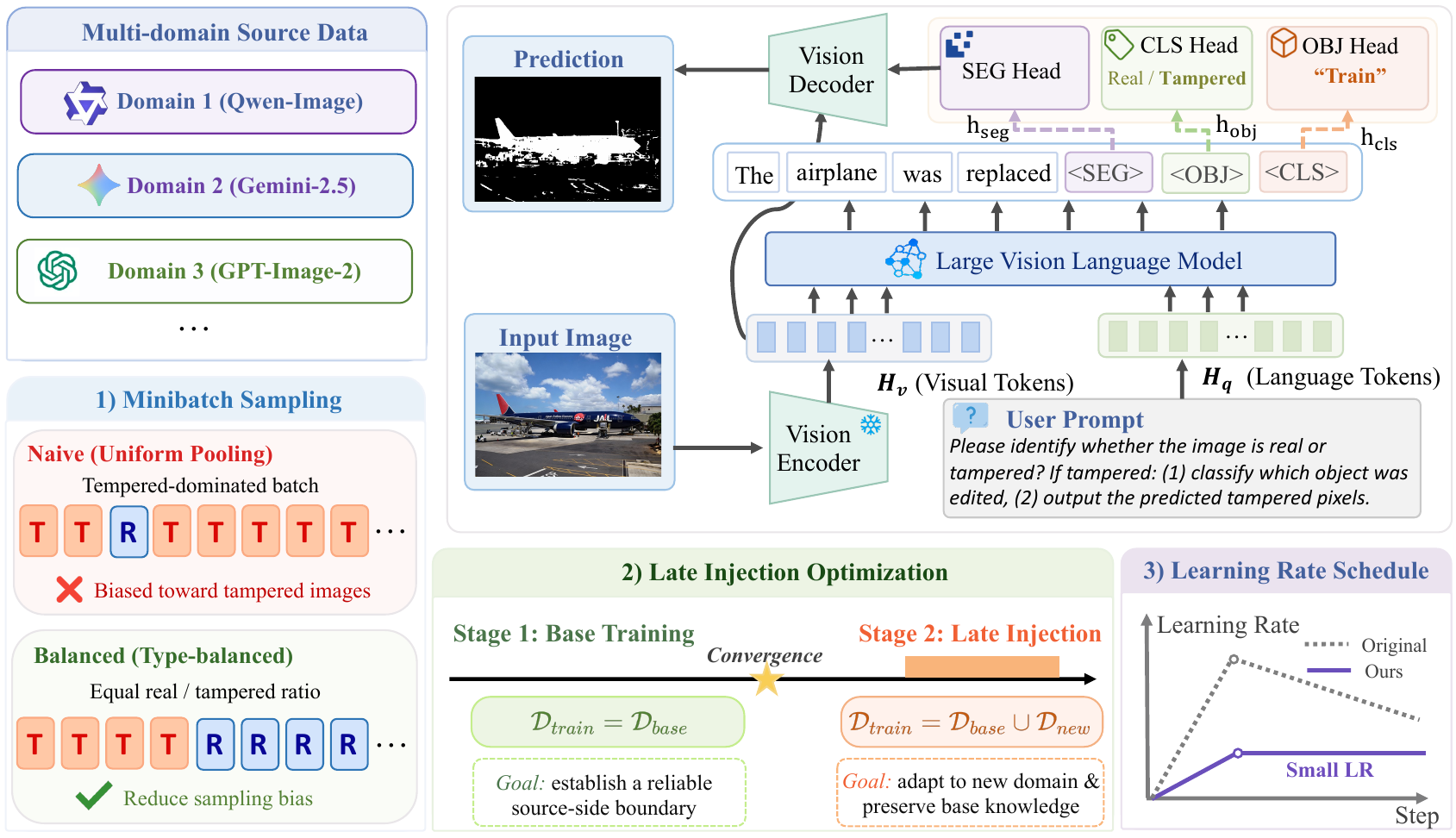}
    \vspace{-15pt}
   \caption{\textbf{Overview of the proposed training framework.}
   Our tamper detector jointly produces a pixel-level tampering mask, a semantic category label, and a natural-language description of the tampered content.
   The training framework integrates three components: 1) balanced real-vs-tampered mini-batch sampling, 2) a late-injection mechanism for incorporating Gemini-2.5 during training, and 3) a low-learning-rate adaptation schedule for pixel-level localization.}
    \label{fig:method}
    \vspace{-12pt}
\end{figure}

\section{Related Work}

\noindent\textbf{Pixel-Level Image Tampering Detection.}
Image tampering detection has evolved from binary image-level classification~\cite{CNNDetection,RECCE,LGrad} to fine-grained localization of edited regions~\cite{chang2023antifakeprompt}. Recent VLM-based methods, such as SIDA~\cite{huang2025sida}, adapt multimodal models for synthetic forgery detection and localization, but still rely on coarse masks that may not faithfully reflect the true editing footprint. PIXAR~\cite{shang2026masks} addresses this limitation by deriving supervision from per-pixel differences between original and edited image pairs, enabling pixel-grounded and semantic-aware tampering detection. However, its training is restricted to a single generator, Qwen-Image~\cite{wu2025qwen}, leaving cross-generator robustness under modern VLMs largely unexplored. In contrast, our work targets generator-agnostic pixel-level tampering detection by explicitly studying domain generalization across diverse and evolving VLM editing distributions.

\noindent\textbf{Domain Generalization.}
Domain generalization~\cite{zhou2022domain,liu2021towards} aims to learn from source domains and generalize to unseen target distributions without target supervision~\cite{gulrajani2020search}. Existing methods improve robustness through controlled adaptation of pre-trained models~\cite{nam2024lipsum}, invariant-representation regularization~\cite{cha2022domain}, or data-balancing strategies~\cite{idrissi2022simple}. However, most are designed for image-level classification and do not directly address pixel-level tampering localization, where local manipulation cues must be preserved under generator shifts. Related generator-agnostic synthetic image detectors exploit up-sampling artifacts~\cite{tan2024rethinking}, frequency cues~\cite{tan2024frequency}, diffusion reconstruction signals~\cite{chen2024drct}, or pre-trained vision-language features~\cite{UnivFD}, but mainly focus on image-level real-vs-generated classification. Our work instead treats VLM generators and editing pipelines as evolving domains and studies cross-generator generalization for pixel-level tampering localization.

\section{Our Approach}
\label{sec:method}

\subsection{Overall Training Framework}

Following the base architecture~\cite{shang2026masks}, our tamper detector $f_\theta$ predicts three complementary outputs: a pixel-level tamper logit map $\mathbf{S}\in\mathbb{R}^{H\times W}$ with probabilities $\widehat{\mathbf{M}}=\sigma(\mathbf{S})$, a multi-label semantic logit vector $\mathbf{z}\in\mathbb{R}^{|\mathcal{C}|}$ with $\hat{\mathbf{y}}=\sigma(\mathbf{z})$, and a natural-language description of the tampered content. As shown in \figref{fig:method}, different task heads are built on hidden features $\mathbf{h}$ from the shared backbone. We optimize the model with five losses.

For semantic prediction, since multiple objects may be tampered in one image, we use sigmoid cross-entropy:
\begin{equation}
\mathcal{L}_\text{sem}
=-\frac{1}{|\mathcal{C}|}\sum_{c\in\mathcal{C}}
\left[y_c\log\hat{y}_c+(1-y_c)\log(1-\hat{y}_c)\right],
\end{equation}
where $y_c$ is the ground-truth semantic label for class $c$. For pixel-level localization, we supervise $\widehat{\mathbf{M}}$ using the thresholded mask $\mathbf{M}_\tau$ with pixel-wise BCE:
\begin{equation}
\mathcal{L}_\text{bce}
=-\frac{1}{HW}\sum_{i,j}
\left[
\mathbf{M}_\tau(i,j)\log\widehat{\mathbf{M}}_{ij}
+(1-\mathbf{M}_\tau(i,j))\log(1-\widehat{\mathbf{M}}_{ij})
\right].
\end{equation}
We further add a DICE loss~\cite{sudre2017generalised} to improve mask quality:
\begin{equation}
\mathcal{L}_\text{dice}
=1-\frac{
2\sum_{i,j}\widehat{\mathbf{M}}_{ij}\mathbf{M}_\tau(i,j)+\varepsilon
}{
\sum_{i,j}\widehat{\mathbf{M}}_{ij}
+\sum_{i,j}\mathbf{M}_\tau(i,j)+\varepsilon
},
\end{equation}
where $\varepsilon$ ensures numerical stability.

To distinguish real and tampered images globally, we use a $\langle\mathrm{CLS}\rangle$-based detection head. Given the last hidden states $\mathbf{H}^{\mathrm{hid}}\in\mathbb{R}^{N\times d}$, we extract $h_{\mathrm{cls}}=\mathbf{H}^{\mathrm{hid}}[\mathrm{CLS}]$ and compute:
\begin{equation}
\mathbf{u}=F_{\mathrm{cls}}(h_{\mathrm{cls}})\in\mathbb{R}^2,\qquad
\hat{\mathbf{p}}=\mathrm{softmax}(\mathbf{u}),\qquad
\mathcal{L}_\text{cls}=\mathcal{L}_{\mathrm{CE}}(\hat{\mathbf{p}},\mathbf{d}),
\end{equation}
where $\mathbf{d}$ is the one-hot label over ${\mathrm{real},\mathrm{tampered}}$. For explanation generation, we train a multimodal causal language model conditioned on the image $\mathbf{I}$ and prompt $P$:
\begin{equation}
\mathcal{L}_{\text{text}}
=-\sum_{i=1}^{L}
\log p_\phi(t_i^*\mid t_{<i}^*,\mathbf{I},P),
\end{equation}
where $T^*=(t_1^*,\dots,t_L^*)$ is the target tamper description.

The final objective combines all task losses:
\begin{equation}
\mathcal{L}_\text{total}
=
\lambda_\text{sem}\mathcal{L}_\text{sem}
+\lambda_\text{bce}\mathcal{L}_\text{bce}
+\lambda_\text{dice}\mathcal{L}_\text{dice}
+\lambda_\text{cls}\mathcal{L}_\text{cls}
+\lambda_\text{text}\mathcal{L}_\text{text}.
\end{equation}
where the $\lambda$ terms balance semantic recognition, pixel localization, image-level detection, and language explanation. Unless otherwise stated, we use $\tau=0.05$ to construct $\mathbf{M}_\tau$ following PIXAR~\cite{shang2026masks}.

\subsection{MiniBatch Sampling}

A key challenge in domain-generalized pixel-level tampering detection is that the training data are usually highly imbalanced across both domains and image types. Let the multi-domain training set be denoted as:
\begin{equation}
    \mathcal{D} =\bigcup_{k=1}^{K} \mathcal{D}^{(k)},
\end{equation}
where $\mathcal{D}^{(k)}$ is the $k$-th domain, corresponding to images generated or edited by a specific VLM, with a unique editing pipeline or data distribution. Each domain contains both real and tampered images:
\begin{equation}
    \mathcal{D}^{(k)} = \mathcal{D}^{(k)}_{\mathrm{real}}
    \cup
    \mathcal{D}^{(k)}_{\mathrm{tamp}}, \qquad \mathcal{D}^{(k)}_{\mathrm{real}}
    \cap
    \mathcal{D}^{(k)}_{\mathrm{tamp}} =\emptyset.
\end{equation}
For each sample $(\mathbf{I}, \mathbf{M}, \mathbf{y}, \mathbf{d}, \texttt{Text}) \in \mathcal{D}$, $\mathbf{I}$ is the input image, $\mathbf{M}$ is the pixel-level tampering mask, $\mathbf{y}$ denotes semantic tampering labels, $\mathbf{d}\in\{\mathrm{real},\mathrm{tampered}\}$ is the image-level label, and $\texttt{Text}$ is the optional tamper description.

A naive sampling strategy draws a minibatch $\mathcal{B}$ from the pooled dataset:
\begin{equation}
    \mathcal{B}  \sim  \mathrm{Uniform}(\mathcal{D}_\mathbf{d}).
\end{equation}
However, this makes the empirical minibatch distribution proportional to the original dataset size of {\em real} and {\em tamp}:
\begin{equation}
    p_{\mathrm{naive}}(\mathrm{real}\  or  \ \mathrm{tamp})  =
    \frac{|\mathcal{D}_\mathbf{d}^{(k)}|}{|\mathcal{D}_\mathrm{real}^{(k)}|+|\mathcal{D}_\mathrm{tamp}^{(k)}|}.
\end{equation}
When real and tampered images are imbalanced, the model may overfit either clean-image priors or tampering-specific artifacts, for instance, the real set is usually much smaller than the tampered one, optimization is dominated by tampered images. To reduce this bias, we construct each minibatch with explicit balancing across image type.
Given a minibatch size $B$, we instead sample a fixed ratio of real and tampered images:
\begin{equation}
    B_{\mathrm{real}}  = \left\lfloor \rho B \right\rfloor, \qquad B_{\mathrm{tamp}}  = B-B_{\mathrm{real}},
\end{equation}
where $\rho\in(0,1)$ controls the real-image ratio. 
This sampler induces the effective training distribution
\begin{equation}
    p_{\mathrm{bal}}(c)  =  \begin{cases}   \rho, & c=\mathrm{real},\\  1-\rho, & c=\mathrm{tamp},  \end{cases}
\end{equation}
Thus, every optimization step receives controllable positive and negative supervision for both global image-level detection and pixel-level localization.

\subsection{Late Injection Optimization}

In practical VLM tampering detection, newly emerging domains\footnote{These domains consist of images generated by newly released advanced VLMs.} are often much smaller than the existing base training set. Directly mixing these scarce samples from the beginning may bias representation learning or cause the model to overfit to unstable domain-specific artifacts. We therefore adopt a simple \emph{late injection} strategy: the detector is first trained on sufficient base-domain data until convergence, and the small new-domain data are injected only in the later training stage.

Let the training data be divided into a large base set and a small new-domain set:
\begin{equation}
   \mathcal{D}  =  \mathcal{D}_{\mathrm{base}}  \cup   \mathcal{D}_{\mathrm{new}},   \qquad  |\mathcal{D}_{\mathrm{base}}|  \gg    |\mathcal{D}_{\mathrm{new}}|.
\end{equation}
The standard joint-training objective is:
\begin{equation}
    \min_{\theta}  \; \mathcal{R}_{\mathrm{joint}}(\theta) = \mathbb{E}_{\mathbf{x}\sim \mathcal{D}_{\mathrm{base}}\cup \mathcal{D}_{\mathrm{new}}}  \left[\ell(f_\theta;\mathbf{x}) \right],
\end{equation}
which uses all domains from the beginning. Instead, we optimize the model in two stages.

In the first stage, we train only on the base-domain data:
\begin{equation}
    \theta_{t+1}    =   \theta_t   -   \eta_t  \nabla_\theta   \mathcal{R}_{\mathrm{base}}(\theta_t),   \qquad  t < T_{\mathrm{inj}},
\end{equation}
where
\begin{equation}
    \mathcal{R}_{\mathrm{base}}(\theta)  =  \mathbb{E}_{\mathbf{x}\sim \mathcal{D}_{\mathrm{base}}}
    \left[\ell(f_\theta;\mathbf{x})\right],
\end{equation}
and $T_{\mathrm{inj}}$ denotes the injection step. This stage allows the detector to learn stable and transferable pixel-level tampering representations from sufficiently diverse base data.

After the base model converges, we inject the new-domain data and continue training on the mixed distribution:
\begin{equation}
    \mathcal{D}_{\mathrm{mix}} = \mathcal{D}_{\mathrm{base}} \cup   \mathcal{D}_{\mathrm{new}}, \qquad  t \geq T_{\mathrm{inj}}.
\end{equation}
The late-stage objective becomes:
\begin{equation}
    \mathcal{R}_{\mathrm{late}}(\theta)   =  (1-\alpha)  \mathbb{E}_{\mathbf{x}\sim \mathcal{D}_{\mathrm{base}}}   \left[\ell(f_\theta;\mathbf{x}) \right]   +  \alpha  \mathbb{E}_{\mathbf{x}\sim \mathcal{D}_{\mathrm{new}}}  \left[\ell(f_\theta;\mathbf{x})\right],
\end{equation}
where $\alpha\in[0,1]$ controls the contribution of the new-domain distribution. 
The parameter update is then:
\begin{equation}
    \theta_{t+1}   =  \theta_t  -  \eta_{\mathrm{late}}  \nabla_\theta   \mathcal{R}_{\mathrm{late}}(\theta_t),  \qquad   t \geq T_{\mathrm{inj}}.
\end{equation}
Equivalently, the training distribution can be written as a time-dependent mixture:
\begin{equation}
    p_t(\mathbf{x})    =  \begin{cases} p_{\mathrm{base}}(\mathbf{x}), & t<T_{\mathrm{inj}},\\  (1-\alpha)p_{\mathrm{base}}(\mathbf{x})+\alpha p_{\mathrm{new}}(\mathbf{x}), & t\geq T_{\mathrm{inj}}.
    \end{cases}
\end{equation}
This formulation prevents the small new-domain set from dominating early feature learning, while still allowing the converged base detector to absorb useful signals from the emerging distribution.

\subsection{Other Practical Training Tricks and Recipes}

We mainly follow the original PIXAR training protocol and modify the learning rate and scheduler to improve training stability under our domain-generalized setting. Specifically, we reduce the base learning rate from $1\times10^{-4}$ to $2\times10^{-5}$, which prevents aggressive parameter updates and helps preserve transferable pixel-level cues learned from multi-domain data. We also replace the original {\texttt{warmup\_decay}} scheduler with a constant scheduler. In the original setting, the learning rate is linearly warmed up from $0$ to $1\times10^{-4}$ during the first $100$ steps, and then linearly decayed from step $100$ to the final training step. In contrast, our constant schedule keeps the learning rate at $2\times10^{-5}$ after the warm-up stage, providing a more conservative optimization trajectory that reduces overfitting to specific source domains or newly injected small distributions. This minor modification is simple but effective for improving cross-domain robustness and stable pixel-level localization.

\section{Experiments}
\label{sec:experiments}

We conduct extensive experiments to demonstrate the efficacy of \ours, with the experimental setup detailed in~\secref{sec:setup}, comparisons against state-of-the-art methods in~\secref{sec:exp_sota}, analysis of training source selection in ~\secref{sec:exp_ablation_data}, and ablation studies with visualization in~\secref{sec:exp_ablation}.

\subsection{Experimental Setup}
\label{sec:setup}

\paragraph{Training and Test Dataset Details.}
We build our training and test sets on PIXAR benchmark~\citep{shang2026masks}.
PIXAR contains a training set of 380K tampered images generated by Qwen-Image VLMs~\cite{wu2025qwen}, and a test set produced by six generators: Qwen-Image, GPT-Image-1.5~\citep{gptimage15}, Gemini-3~\citep{gemini3}, Flux.2~\citep{flux2}, Seedream-4.5~\citep{seedream2025seedream40}, and Gemini-2.5~\citep{gemini25}. 
Additional details of PIXAR are provided in~\appref{app_sec:pixar}.

We construct our training set from two domain sources.
The base source is a $70\text{K}$ subset of Qwen-Image tampered images sampled from the PIXAR training set. 
We choose Qwen-Image because it is an open-source generator and, among all candidate base sources, yields the strongest cross-generator generalization (see \secref{sec:exp_ablation_data}).
To further improve robustness to generator shift, we add only 3K Gemini-2.5~\citep{gemini25} tampered images from a held-out partition of the PIXAR Gemini-2.5 test pool \footnote{These Gemini-2.5 samples are used exclusively for training and are removed from our final evaluation set.}. 
Gemini-2.5 is selected as the new domain source because it provides the largest OOD improvement among all candidate companion generators (see \secref{sec:exp_ablation_data}). 
Thus, our final training set contains $73\text{K}$ tampered images, amounting to only $19.2\%$ of the original PIXAR training scale, while retaining strong generalization to unseen generators.
Unless otherwise stated, we set $\rho=0.5$, so each mini-batch contains the same number of real and tampered images.

For the test set, we use the remaining PIXAR multi-generator test data after excluding the Gemini-2.5 training subset.
To better evaluate generalization under recent generators, we adapt the PIXAR test protocol by excluding the GPT-Image-1.5 and Gemini-3 subsets and incorporating samples from newer generators, including GPT-Image-2.0~\citep{openai2026gptimage2} and Gemini-3.1~\citep{googledeepmind2026gemini31procard}. 
Under this protocol, Qwen-Image~\cite{wu2025qwen} and Gemini-2.5~\cite{gemini25} act as the in-domain (ID) generators, while the remaining four models, GPT-Image-2.0~\cite{openai2026gptimage2}, Gemini-3.1~\cite{googledeepmind2026gemini31procard}, FLUX.2~\cite{flux2}, and Seedream 4.5~\cite{seedream2025seedream40}, are held out and treated as out-of-distribution (OOD) generators unseen during training.
Detailed statistics of our final test distribution are reported in \tabref{tab:app_test_distribution} in \appref{app_sec:test_distribution}.

\vspace{3pt}
\noindent{\bf Baselines and Metrics.}
We compare against three strong baselines, including (1) PIXAR~\citep{shang2026masks}, the current state of the art pixel-level detection method; (2) SIDA~\citep{huang2025sida}, and (3) LISA~\citep{lai2024lisa}. For each baseline, we include both 7B and 13B variants.
Following the evaluation protocol of PIXAR~\citep{shang2026masks}, we assess pixel-level localization using g-IoU and c-IoU for overall pixel-wise accuracy.
We also report binary real-vs-tampered classification accuracy.
To explicitly assess cross-generator generalization, we additionally evaluate two generator-level metrics: ID average accuracy and OOD average accuracy.

\vspace{3pt}
\noindent{\bf Implementation Details.}
We employ parameter-efficient fine-tuning via LoRA \citep{hu2022lora} with rank $r=8$, scaling factor $\alpha=16$, and a dropout rate of $0.05$. 
During training, the CLIP vision tower, multimodal projector, and SAM image encoder remain frozen, while the SAM mask decoder and the four task-specific heads are fine-tuned.
All configurations are trained for $5$ epochs, with each epoch consisting of $500$ optimization steps. 
We use a batch size of $8$ and the AdamW optimizer \citep{loshchilov2017decoupled} with $(\beta_1, \beta_2) = (0.9, 0.95)$ and zero weight decay. 
All experiments are conducted on 4 RTX A6000 GPUs, and results are reported across three independent random seeds to ensure statistical robustness.

\begin{table}[t]
    \caption{\textbf{Pixel-level localization results on the four OOD generators.} ``Avg.'' denotes the average results across the four generators. \textbf{Bold} denotes the best result, and \underline{underline} denotes the second-best result.}
    \label{tab:main_results_pixel}
    \centering
    \renewcommand{\arraystretch}{1.1}
    \resizebox{\linewidth}{!}{
    \begin{tabular}{lcc|ccccc|ccccc}
        \toprule
        \multicolumn{1}{c}{\multirow{2}{*}{Method}} &
        \multicolumn{1}{c}{\multirow{2}{*}{Pixel Recall}} &
        \multicolumn{1}{c|}{\multirow{2}{*}{Pixel F1}} &
        \multicolumn{5}{c|}{Per-generator gIoU} &
        \multicolumn{5}{c}{Per-generator cIoU} \\
        \cmidrule(lr){4-8}
        \cmidrule(lr){9-13}
        & & &
        \multicolumn{1}{c}{GPT-Image-2.0} &
        \multicolumn{1}{c}{Gemini-3.1} &
        \multicolumn{1}{c}{FLUX.2} &
        \multicolumn{1}{c}{Seedream 4.5} &
        \multicolumn{1}{c|}{Avg.} &
        \multicolumn{1}{c}{GPT-Image-2.0} &
        \multicolumn{1}{c}{Gemini-3.1} &
        \multicolumn{1}{c}{FLUX.2} &
        \multicolumn{1}{c}{Seedream 4.5} &
        \multicolumn{1}{c}{Avg.} \\
        \midrule
        LISA-7B~\cite{lai2024lisa}    & 1.30 & 2.54 & 0.014 & 0.018 & 0.009 & 0.012 & 0.013 & 0.015 & 0.021 & 0.007 & 0.012 & 0.014 \\
        SIDA-7B~\cite{huang2025sida}    & \underline{33.71} & 27.42 & 0.147 & \underline{0.164} & \underline{0.186} & \underline{0.155} & \underline{0.163} & 0.145 & 0.156 & \underline{0.194} & 0.142 & 0.159 \\
        PIXAR-7B~\cite{shang2026masks}    & 25.77 & \underline{28.49} & \underline{0.155} & 0.161 & 0.170 & 0.150 & 0.159 & \underline{0.159} & \underline{0.161} & 0.184 & \underline{0.161} & \underline{0.166} \\
        \rowcolor[HTML]{F2F2F2}
       \bf \ours-7B & \textbf{45.11} & \textbf{33.29} & \textbf{0.189} & \textbf{0.195} & \textbf{0.222} & \textbf{0.166} & \textbf{0.193} & \textbf{0.199} & \textbf{0.187} & \textbf{0.243} & \textbf{0.176} & \textbf{0.201} \\
        \midrule\midrule
        LISA-13B~\cite{lai2024lisa}    & 3.06 & 5.73 & 0.026 & 0.038 & 0.021 & 0.029 & 0.028 & 0.028 & 0.045 & 0.020 & 0.032 & 0.031 \\
        SIDA-13B~\cite{huang2025sida}    & 16.84 & 21.25 & 0.110 & 0.134 & 0.119 & 0.116 & 0.120 & 0.113 & 0.136 & 0.117 & 0.109 & 0.119 \\
        PIXAR-13B~\cite{shang2026masks}    & \underline{33.49} & \underline{30.95} & \underline{0.177} & \underline{0.188} & \underline{0.182} & \underline{0.157} & \underline{0.176} & \underline{0.184} & \underline{0.185} & \underline{0.194} & \underline{0.170} & \underline{0.183} \\
        \rowcolor[HTML]{F2F2F2}
       \bf \ours-13B & \textbf{62.19} & \textbf{37.42} & \textbf{0.201} & \textbf{0.198} & \textbf{0.293} & \textbf{0.197} & \textbf{0.222} & \textbf{0.210} & \textbf{0.195} & \textbf{0.318} & \textbf{0.206} & \textbf{0.232} \\
        \bottomrule
    \end{tabular}
    }
    \vspace{-0.1in}
\end{table}

\subsection{Comparison with SOTA Methods}
\label{sec:exp_sota}
We report the cross-generator pixel-level localization performance across the four held-out out-of-domain generators in~\tabref{tab:main_results_pixel}.
It is evident that \ours \textit{consistently achieves state-of-the-art performance}, demonstrating robust zero-shot transferability to unseen generators.
Across all evaluation metrics, \ours substantially outperforms the strongest prior baseline PIXAR-7B~\citep{shang2026masks}, with absolute gains of $19.34\%$ in Pixel Recall and $4.80\%$ in Pixel F1, a $21.4\%$ relative improvement in average gIoU, and a $21.1\%$ relative improvement in average cIoU.
It is worth noting that \ours uses substantially less training data than PIXAR. 
While PIXAR is trained on the full $380\text{K}$ Qwen-Image training set, \ours uses only $73\text{K}$ tampered images, corresponding to $19.2\%$ of the original training scale.
This suggests that simply increasing the amount of same-generator data is insufficient for cross-generator robustness; the diversity of the training sources and the stability of the optimization process also play a critical role.
At the 13B scale, \ours-13B remains the best on every averaged metric, lifting average gIoU from $0.176$ to $0.222$ and average cIoU from $0.183$ to $0.232$ over PIXAR-13B, while also yielding a $6.47\%$ gains in Pixel F1.

\begin{wraptable}{r}{0.6\columnwidth}
    \vspace{-12pt}
    \caption{Binary classification results on the four OOD generators.  
  \textbf{Bold} denotes the best result.
    }
    \vspace{-5pt}
    \label{tab:main_results_binary}
    \centering
    \renewcommand{\arraystretch}{1}
    \resizebox{1\linewidth}{!}{  
        \begin{tabular}{lccccc}
        \toprule
        \multicolumn{1}{c}{\multirow{1}{*}{Method}} & \multicolumn{5}{c}{Per-generator Accuracy (\%)} \\
        \cmidrule(lr){2-6}
        & \multicolumn{1}{c}{GPT-Image-2.0} & \multicolumn{1}{c}{Gemini-3.1} & \multicolumn{1}{c}{FLUX.2} & \multicolumn{1}{c}{Seedream 4.5} & \multicolumn{1}{c}{Avg.} \\
        \midrule
        SIDA-7B~\cite{huang2025sida}    & 13.7 & 23.2 & 15.2 & 20.5 & 18.1 \\
        PIXAR-7B~\cite{shang2026masks}    & \underline{78.0} & \underline{62.0} & \underline{55.1} & \underline{83.2} & \underline{69.6} \\
        \rowcolor[HTML]{F2F2F2} 
        \bf \ours-7B   & \textbf{84.9} & \textbf{84.0} & \textbf{61.3} & \textbf{88.3} & \textbf{79.6} \\
        \midrule
        SIDA-13B~\cite{huang2025sida}    & 20.0 & 38.5 & 21.9 & 38.6 & 29.8 \\
        PIXAR-13B~\cite{shang2026masks}    & \underline{67.5} & \underline{46.0} & \underline{46.9} & \underline{75.4} & \underline{59.0} \\
        \rowcolor[HTML]{F2F2F2}
        \bf \ours-13B   & \textbf{88.4} & \textbf{77.6} & \textbf{78.1} & \textbf{92.0} & \textbf{84.0} \\
        \bottomrule
    \end{tabular}
        }
    \vspace{-5pt}
\end{wraptable}

We also report per-generator results, demonstrating that \ours improves localization on all OOD generators.
The largest absolute gain at 7B scale is observed on FLUX.2, where gIoU increases from $0.170$ to $0.222$, achieving $30.6\%$ relative gains.
These results indicate that introducing only a small companion source, Gemini-2.5, \textit{generalizes far beyond Gemini-2.5 itself and translates into broad cross-generator robustness}.
We further report binary real-vs-tampered classification results in~\tabref{tab:main_results_binary}.
\ours achieves the best average per-generator OOD accuracy at both 7B and 13B scales, with absolute gains of $10.0\%$ over PIXAR-7B and $25.0\%$ over PIXAR-13B, respectively.
Additional in-domain results are reported in~\tabref{tab:main_results_id} of \appref{app_sec:in_domain}.

\begin{table}[t]
    \caption{\textbf{Per-generator cross-domain generalization ability.} Each row corresponds to a single-generator training configuration, and each of the first six numeric columns reports the tampered accuracy (\%) on the test split of the corresponding generator. Diagonal values indicate \hltext{in-domain} performance; off-diagonal entries are OOD. ``Avg.'' denotes the mean of the five off-diagonal entries in each row, respectively; ``Worst'' denotes the worst-performing test generator, corresponding to the values in \hlBtext{Worst}.}
    \label{tab:transfer-matrix}
    \centering
    \renewcommand{\arraystretch}{1.1}
    \resizebox{\linewidth}{!}{
    \begin{tabular}{l|cccccc|cc}
        \toprule
        \multirow{2}{*}{Training domain} & \multicolumn{6}{c|}{Test generator Accuracy (\%)} & \multicolumn{2}{c}{OOD} \\
        \cmidrule(lr){2-7}\cmidrule(lr){8-9}
        & Qwen-Image & GPT-Image-2.0 & Gemini-3.1 & FLUX.2 & Seedream 4.5 & Gemini-2.5 & OOD Avg. & Worst \\
        \midrule
        Qwen-Image      & \hl{\textbf{82.7}} & 69.8 & 52.4 & 56.3 & 83.8 & \hlB{41.2} & \textbf{60.7} & Gemini-2.5 \\
        GPT-Image-2.0   & 14.8 & \hl{\textbf{45.5}} & \hlB{\phantom{0}0.8} & 21.3 & 20.3 & \phantom{0}2.6 & 11.9 & Gemini-3.1 \\
        Gemini-3.1        & 10.2 & \hlB{\phantom{0}3.0} & \hl{\textbf{72.5}} & \phantom{0}4.2 & 13.5 & 38.5 & 13.9 & GPT-Image-2.0 \\
        FLUX.2          & 24.4 & 30.2 & \hlB{\phantom{0}2.6} & \hl{\textbf{85.5}} & 17.9 & 10.3 & 17.1 & Gemini-3.1 \\
        Seedream 4.5    & 47.2 & 29.4 & 17.4 & 21.3 & \hl{\textbf{70.8}} & \hlB{14.5} & 26.0 & Gemini-2.5 \\
        Gemini-2.5      & 23.0 & 18.9 & 62.1 & \hlB{12.5} & 25.0 & \hl{\textbf{59.0}} & 28.3 & FLUX.2 \\
        \bottomrule
    \end{tabular}
    }
    \vspace{-10pt}
\end{table}

\vspace{-5pt}
\subsection{Why Qwen and Gemini-2.5 as Training Data Sources?}
\label{sec:exp_ablation_data}

\noindent{\bf Qwen-Image as the Base Source Generator.}
We choose Qwen-Image as our base source generator for two reasons.
First, Qwen-Image is open-source and supported by the largest-scale training pool in PIXAR ($380\text{K}$ samples), making it the most practical candidate for large-scale supervision.
Second, and more importantly, we empirically verify that Qwen-Image yields the strongest cross-generator generalization among all six candidate generators.
Specifically, we train six single-source variants, each fine-tuning the detector on tampered images from one generator alone, and evaluate every variant on a fixed test set covering all six generators: the diagonal entries report in-domain accuracy on the training generator, while the five off-diagonal entries report OOD accuracy on the unseen generators. 
All variants use the identical training budgets, optimization schedules, and architectural settings.

The results of the cross-domain generalization ability of each single-generator training source are reported in \tabref{tab:transfer-matrix}.
It is evident that \textit{Qwen-Image achieves the highest average accuracy of $60.7\%$}, surpassing the second-best source, Gemini-2.5, by $32.4\%$, while also attaining the strongest worst-case OOD accuracy ($41.2\%$ on Gemini-2.5) by a wide margin: every other training source has a substantially lower worst-case OOD accuracy (all $\leq 14.5\%$), with several collapsing to single digits (e.g., $2.6\%$ for GPT-Image-2.0-trained models on Gemini-2.5, and $2.6\%$ for FLUX.2-trained models on Gemini-3.1).
This combination of the highest mean and the highest worst-case accuracy indicates that Qwen-Image yields detectors that generalize both broadly and robustly across unseen generators, in contrast to other sources whose generalization is sharply uneven.
We therefore fix Qwen-Image as the base source in all experiments.

\noindent{\bf Gemini-2.5 as the Optimal Companion Source.}
Having fixed Qwen-Image as the base source, we now investigate which additional source should be combined with it to maximize OOD generalization.
\tabref{tab:source_composition} reports the results of combining Qwen-Image with each remaining generator individually, while results for compositions with more sources are provided in~\appref{app_sec:more_source_composition}.

As shown in \tabref{tab:source_composition}, \textit{Qwen-Image + Gemini-2.5 achieves the highest Avg.~(OOD) of $66.3\%$}, outperforming all four alternative companion choices by a clear margin: the Gemini-3.1 reaches only $63.6\%$, and the remaining choices fall below $60\%$.
This finding is further reinforced by \tabref{tab:transfer-matrix}: Gemini-2.5 is the worst-case OOD target for Qwen-Image-trained detectors, i.e.\ the generator on which the base source generalizes most poorly.
The convergence of these observations motivates our final design: \ours uses \textit{Qwen-Image $+$ Gemini-2.5} as the training sources.

\begin{table}[t]
    \caption{\textbf{Source composition study.} Each row represents a different training source composition built on top of Qwen-Image. Per-generator columns report tampered accuracy (\%) on each test generator. \textit{Avg. (All)} denotes the average accuracy across all six generators; \textit{Avg. (OOD)} denotes the average accuracy across the held-out OOD generators. \textbf{Bold} indicates the best result, and \underline{underline} indicates the second-best result.
    }
    \label{tab:source_composition}
    \centering
    \renewcommand{\arraystretch}{1.1}
    \resizebox{\linewidth}{!}{
    \begin{tabular}{l|cccccc|cc}
        \toprule
        \multirow{2}{*}{Training composition} &
        \multicolumn{6}{c|}{Per-generator Accuracy (\%)} &
        \multicolumn{2}{c}{Average} \\
        \cmidrule(lr){2-7}\cmidrule(lr){8-9}
        & Qwen-Image & GPT-Image-2.0 & Gemini-3.1 & Gemini-2.5 & FLUX.2 & Seedream 4.5 & All & OOD \\
        \midrule
        \rowcolor[HTML]{F2F2F2}
        Qwen-Image + Gemini-2.5
            & \textbf{92.6} & \textbf{71.5} & \underline{56.6} & \textbf{53.0} & \underline{53.2} & \textbf{83.8} & \textbf{68.5} & \textbf{66.3} \\
        Qwen-Image + FLUX.2
            & 90.3 & 70.5 & 45.2 & 40.8 & \textbf{59.2} & 79.7 & 64.3 & 59.1 \\
        Qwen-Image + Seedream 4.5
            & 88.1 & 71.1 & 50.0 & 42.7 & 52.6 & \underline{82.7} & 64.5 & 54.1 \\
        Qwen-Image + Gemini-3.1
            & \underline{90.6} & \textbf{71.5} & \textbf{57.8} & \underline{48.9} & \underline{53.2} & 80.5 & \underline{67.1} & \underline{63.6} \\
        Qwen-Image + GPT-Image-2.0
            & 88.3 & 70.1 & 45.0 & 41.7 & 52.9 & 78.2 & 62.7 & 54.4 \\
        \bottomrule
    \end{tabular}
    }
    \vspace{-10pt}
\end{table}

\begin{wraptable}{r}{0.6\columnwidth}
    \vspace{-12pt}
     \caption{\textbf{Influence of base-source training data size.} We vary the number of Qwen-Image tampered samples $N$ used for training while keeping all other settings identical.}
    \vspace{-5pt}
    \label{tab:qwen_size}
    \centering
    \renewcommand{\arraystretch}{1}
    \resizebox{1\linewidth}{!}{  
        \begin{tabular}{c|cccc|cc}
        \toprule
        \multirow{2}{*}{$N$} & \multicolumn{4}{c|}{Per-generator Accuracy (\%)} & \multicolumn{2}{c}{Average} \\
        \cmidrule(lr){2-5}\cmidrule(lr){6-7}
        & GPT-Image-2.0 & Gemini-3.1 & FLUX.2 & Seedream 4.5 & Acc. & Pixel F1 \\
        \midrule
        30K                                         & 70.6 & \underline{64.4} & 49.0 & 75.8 & \underline{65.0} & \underline{27.20} \\
        \rowcolor[HTML]{F2F2F2}
        70K
           & \textbf{84.9} & \textbf{84.0} & \textbf{61.3} & \textbf{88.3} & \textbf{79.6} & \textbf{33.40} \\
        150K                                        & 74.8 & 52.6 & 51.7 & \underline{80.3} & 64.9 & 24.63 \\
        380K (full)                                 & \underline{75.5} & 47.1 & \underline{52.0} & 79.1 & 63.4 & 26.74 \\
        \bottomrule
        \end{tabular}
    }
    \vspace{-5pt}
\end{wraptable}

\vspace{3pt}
\noindent\textbf{Influence of Base-Source Data Size.}
A natural question is how much Qwen-Image data is needed when paired with our small companion source ($3\text{K}$ Gemini-2.5). 
To answer this, we vary the number of Qwen-Image tampered samples $N$ from $30\text{K}$ up to the full $380\text{K}$, while keeping every other component of the training recipe fixed. 
Results are reported in \tabref{tab:qwen_size}.
It is evident that increasing the Qwen-Image subset from $30\text{K}$ to $70\text{K}$ improves both average accuracy and Pixel F1.
Beyond this point, however, additional same-source data brings no further gain in cross-generator generalization.
One possible reason is that as the Qwen-Image portion grows, the relatively small Gemini-2.5 supervision constitutes a smaller fraction of each mini-batch, weakening its contribution to OOD generalization.
\textit{We therefore pair $N{=}70$K Qwen-Image samples with $3$K Gemini-2.5 tampered samples in all main experiments, corresponding to only $19.2\%$ of the full PIXAR training pool.}

\subsection{Ablation Study}
\label{sec:exp_ablation}

\begin{wraptable}{r}{0.55\columnwidth}
    \vspace{-12pt}
    \caption{\textbf{Results under the same training data.} All methods are trained on the same training data.}
    \label{tab:same_data}
    \centering
    \renewcommand{\arraystretch}{1}
    \resizebox{1\linewidth}{!}{  
        \begin{tabular}{l|cccc|cc}
        \toprule
        \multirow{2}{*}{Method} & \multicolumn{4}{c|}{Per-generator Accuracy (\%)} & \multicolumn{2}{c}{Average} \\
        \cmidrule(lr){2-5}\cmidrule(lr){6-7}
        & GPT-Image-2.0 & Gemini-3.1 & FLUX.2 & Seedream 4.5 & Acc. & Pixel F1 \\
        \midrule
        SIDA          & 49.0 & 39.0 & 35.8 & \underline{74.3} & 49.5 & 28.37 \\
        PIXAR         & \underline{50.3} & \underline{44.1} & \underline{41.6} & 72.1 & \underline{52.0} & \underline{28.59} \\
        \rowcolor[HTML]{F2F2F2}
        \bf \ours     & \textbf{71.5} & \textbf{56.6} & \textbf{53.2} & \textbf{83.8} & \textbf{66.3} & \textbf{29.89} \\
        \bottomrule
        \end{tabular}
    }
\end{wraptable}

\paragraph{Same Training Data.}
To verify that the improvements of \ours are not solely attributable to the use of multi-source training data, we retrain all baselines on the same Qwen-Image $+$ Gemini-2.5 training data, using identical training budgets and optimization settings as \ours.
Results are reported in~\tabref{tab:same_data}.
\textit{\ours continues to outperform every baseline when trained on the same data}, attaining $66.3\%$ average accuracy and $29.89\%$ Pixel F1, compared to $52.0\%$ accuracy and $28.59\%$ Pixel F1 achieved by the strong baseline PIXAR.
This gap demonstrates that the improvements of \ours arise from the method-level contributions in~\secref{sec:method}, rather than from access to richer training data alone.

\vspace{3pt}
\noindent\textbf{Ablation on Three Training Components.}
We further conduct an ablation study to validate the contribution of each component in our training strategy: \textit{MiniBatch Sampling (MS)}, \textit{Late Injection (LI)}, and the \textit{Learning Rate (LR) Schedule}.
Results are reported in~\tabref{tab:ablation}: \textit{every component can yield improvement, and combining all three achieves the strongest OOD generalization}.

\begin{wraptable}{r}{0.65\columnwidth}
    \caption{\textbf{Ablation study on three training components (pixel-level localization).}
    All variants share the same training data and settings. 
    Per-generator columns report pixel-F1 (\%) on each held-out (OOD) generator.
    }
    \vspace{-5pt}
    \label{tab:ablation}
    \centering
    \renewcommand{\arraystretch}{1}
    \resizebox{1\linewidth}{!}{
        \begin{tabular}{ccc|cccc|cc}
        \toprule
        \multicolumn{3}{c|}{Components} & \multicolumn{4}{c|}{Per-generator Pixel-F1 (\%)} & \multicolumn{2}{c}{Average} \\
        \cmidrule(lr){1-3}\cmidrule(lr){4-7}\cmidrule(lr){8-9}
        MS & LI & LR & GPT-Image-2.0 & Gemini-3.1 & FLUX.2 & Seedream 4.5 & Acc. & Pixel F1 \\
        \midrule
        \checkmark & \xmark     & \xmark     & 28.12 & 27.28 & 26.72 & 23.99 & \underline{65.3} & 26.53 \\
        \checkmark & \checkmark & \xmark     & \underline{30.16} & \underline{28.44} & \underline{30.21} & \underline{25.78} & 64.0 & \underline{28.65} \\
        \checkmark & \checkmark & \checkmark & \textbf{30.46} & \textbf{29.05} & \textbf{32.82} & \textbf{27.21} & \textbf{66.3} & \textbf{29.89} \\
        \bottomrule
        \end{tabular}
    }
    \vspace{-5pt}
\end{wraptable}

\begin{figure}[!t]
    \centering
    \begin{subfigure}[b]{0.36\linewidth}
        \includegraphics[width=\linewidth]{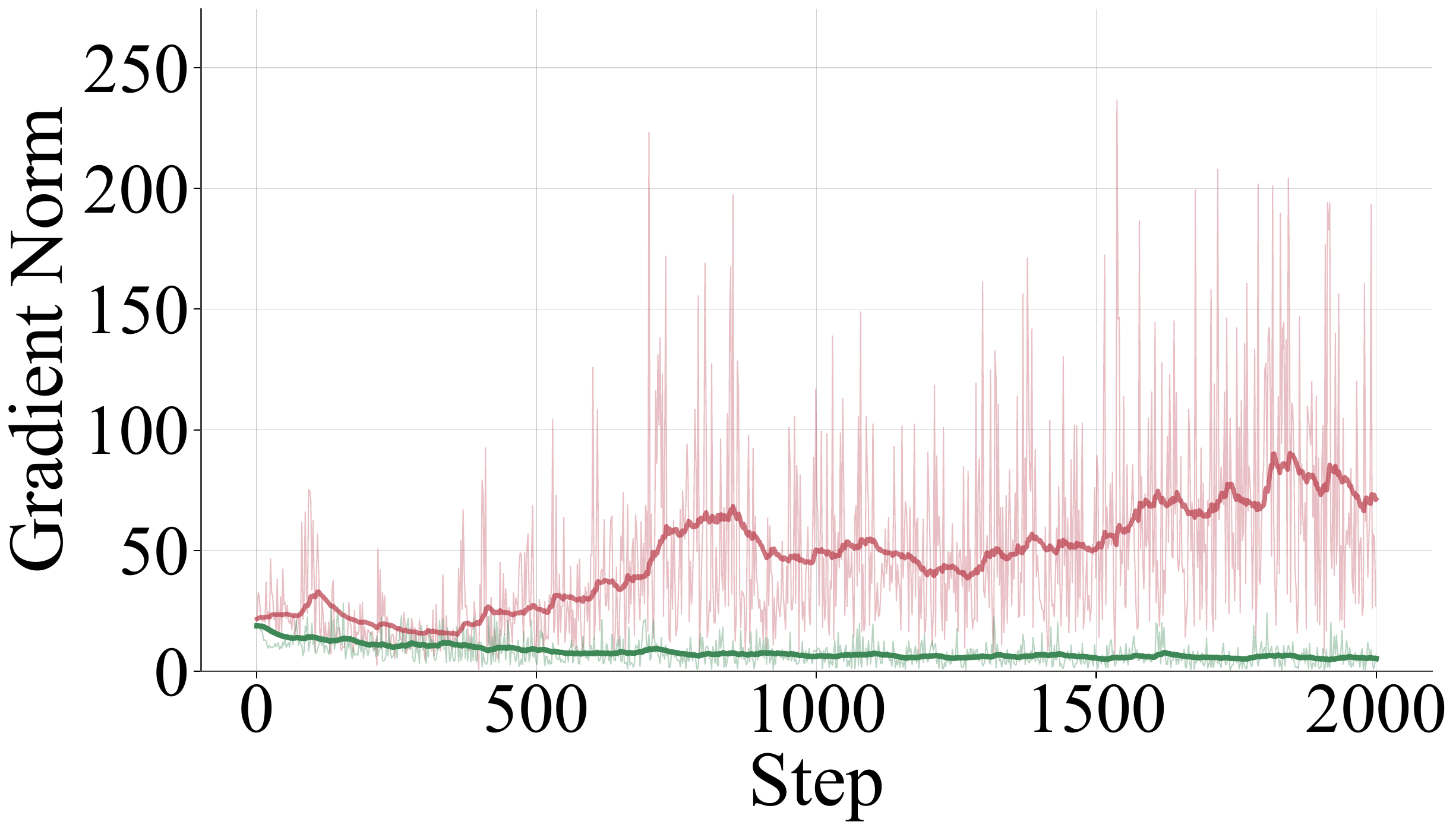}
        \caption{Gradient Norm}
        \label{fig:ablation_grad_norm}
    \end{subfigure}\hfill
    \begin{subfigure}[b]{0.31\linewidth}
        \includegraphics[width=\linewidth]{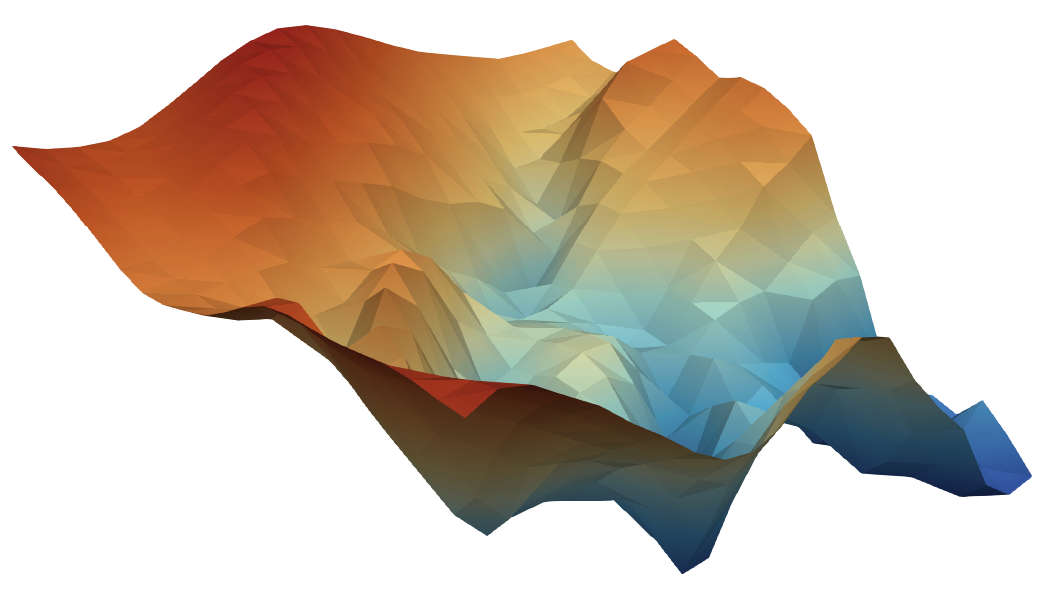}
        \vspace{0.05cm}
        \caption{Loss Landscape (PIXAR \cite{shang2026masks})}
        \label{fig:ablation_loss_baseline}
    \end{subfigure}\hfill
    \begin{subfigure}[b]{0.31\linewidth}
        \includegraphics[width=\linewidth]{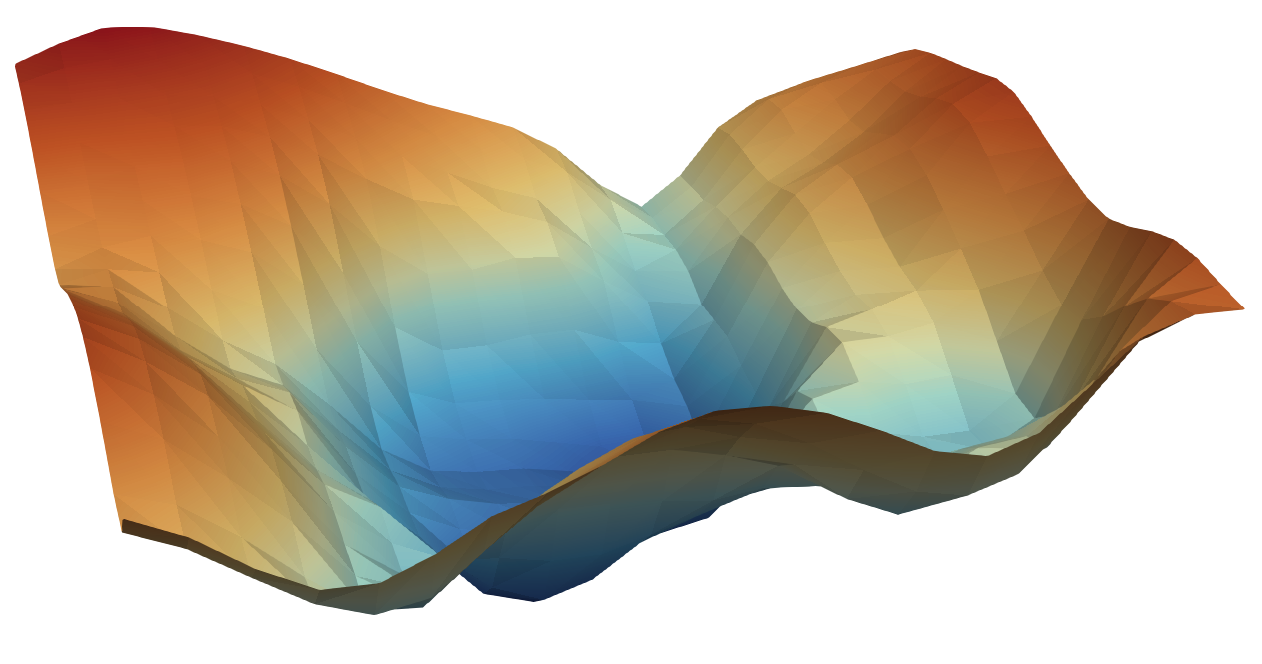}
        \vspace{0.05cm}
        \caption{Loss Landscape (\ours)}
        \label{fig:ablation_loss_ours}
    \end{subfigure}
    \vspace{-0.2cm}
    \caption{\textbf{Influence of mini-batch sampling and learning rate schedule.} (a) Gradient norm of the \texttt{<CLS>} head over training steps: random sampling (\textcolor{darkred}{red}) leads to large fluctuations, while balanced mini-batch sampling (\textcolor{darkgreen}{green}) yields a substantially smoother trajectory. (b) and (c) Loss landscape comparison.}
    \label{fig:ablation_study}
    \vspace{-5pt}
\end{figure}

For \textit{mini-batch sampling}, as shown in~\figref{fig:ablation_grad_norm}, due to the imbalance ratio between real and tampered samples, naive random sampling leads to unstable learning, and the gradient norm of the \texttt{<CLS>} head exhibits large fluctuations (\textcolor{darkred}{red line}).
To address this, we adopt balanced mini-batch sampling, which enforces a fixed real-to-tampered ratio within every mini-batch (more results of different ratios are provided in \tabref{tab:ablation_bt_ratio} of \appref{app_sec:different_ratio}.)
Therefore, the gradient norm becomes substantially smoother and flatter (\textcolor{darkgreen}{green line}).

We further visualize the loss landscape~\citep{li2018visualizing} of \ours with the strongest baseline PIXAR, as shown in ~\figref{fig:ablation_loss_baseline} and~\figref{fig:ablation_loss_ours}.
It is evident that \ours converges to a noticeably flatter minimum than the baseline. This analysis indicates that the simple modules designed in our method play a substantive role in shaping a loss-landscape geometry that benefits OOD generalization.

\begin{figure}[!t]
    \centering
    \includegraphics[width=1\linewidth]{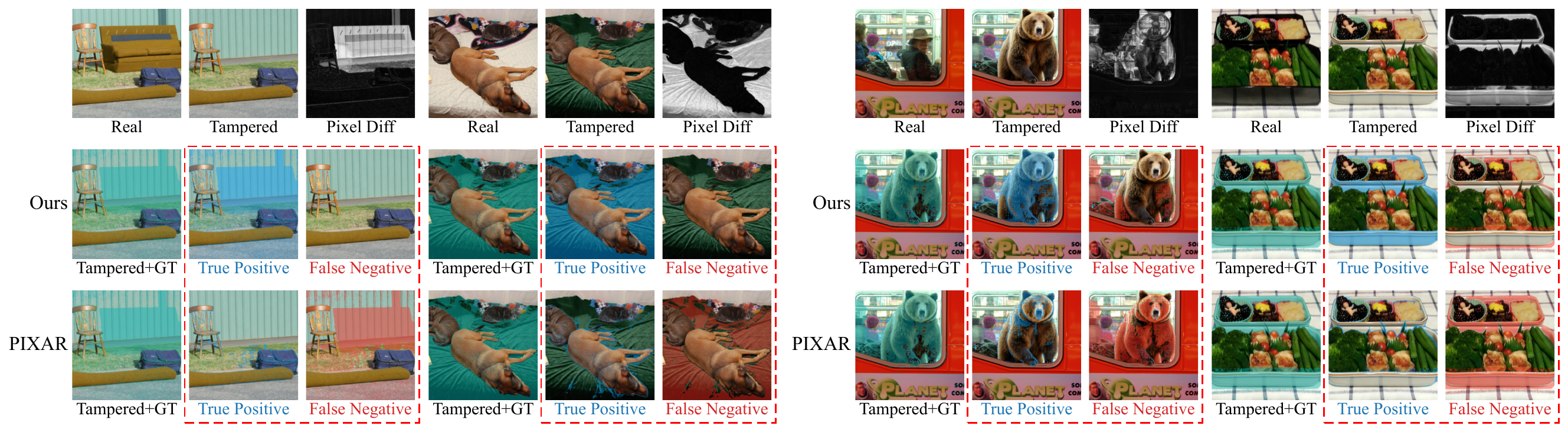}
    \caption{\textbf{Qualitative comparison of predicted tampered pixels between \ours and PIXAR~\citep{shang2026masks}} on the four OOD generators in PIXAR test data (left to right: GPT-Image-2.0, Gemini-3.1, FLUX.2, Seedream-4.5).}
    \label{fig:Visualization}
    \vspace{-15pt}
\end{figure}

\noindent\textbf{Qualitative Visualization.}
To qualitatively evaluate the predictions of \ours against PIXAR, we visualize examples in~\figref{fig:Visualization}, covering all four held-out OOD generators: GPT-Image-2.0, Gemini-3.1, FLUX.2, and Seedream 4.5.
We illustrate both true positives and false negatives.
Across all four OOD generators, \textit{\ours consistently and accurately localizes the truly edited regions.} 
More visualization results are provided in \appref{app_sec:additional_qualitative}.

\vspace{-5pt}
\section{Conclusion}

In this work, we have studied domain generalization for pixel-level image tampering detection in modern VLMs, a new and practical problem for reliable visual authenticity verification. We present a simple yet effective training framework that improves cross-domain robustness without requiring complex architectural changes. By combining balanced real/tampered minibatch sampling, late injection of small emerging domains, and conservative low-learning-rate optimization, our method encourages the detector to learn transferable tampering-related cues rather than overfitting to source-specific artifacts. Extensive experiments show that our approach achieves stronger performance with substantially less training data and consistently improves localization accuracy across new domains. Overall, our framework provides a practical recipe for building robust pixel-level tampering detectors for rapidly evolving VLM-generated and VLM-edited images.

\section*{Acknowledgments}

This work is supported by the United Al Saqer Group Grant and the MBZUAI-WIS Joint Program for Artificial Intelligence Research.


{\small
\bibliographystyle{plain}
\bibliography{resources/main}
}


\newpage
\appendix

\appendix

\section*{{\Large{Appendix}}}

\section{Limitations}

Although our framework improves domain generalization for pixel-level VLM image tampering detection, it still has several limitations. First, its performance depends on the diversity and quality of the available source domains. If the source data does not cover sufficiently representative manipulation patterns, the learned cues may still fail under extreme OOD shifts. Second, late injection requires access to a small amount of labeled data from emerging distributions, which may not always be available for distributions induced by closed-source or newly released VLMs. Third, our method mainly improves the training recipe rather than the detector architecture, so its upper-bound performance remains constrained by the capacity and design of the underlying segmentation-based model.

\section{Societal Impacts}

This work may have positive societal impacts by improving the reliability of visual authenticity verification in the era of modern VLMs. Stronger pixel-level tampering detection results can help identify manipulated regions in images, support fact-checking and journalism, assist content moderation, and provide forensic evidence for detecting misinformation, fraud, and malicious image edits. Because the method is designed for domain generalization, it may remain useful as new VLMs and editing tools emerge, reducing the need to retrain detectors from scratch for every new model.

At the same time, the work may also have negative societal impacts. More capable tampering detectors could be misused to build stronger adversarial attacks, as attackers may study detector behavior to produce edits that are harder to localize. There is also a risk of over-reliance on automated detection systems: false positives may incorrectly flag benign or legitimate images as manipulated, while false negatives may provide a false sense of trust in harmful manipulated content. In addition, if training data are biased toward certain image domains, objects, cultures, or editing styles, the detector may perform unevenly across different populations or contexts. 

\section{More Experimental Details and Results}
\label{app_sec:exp}

Here we summarize the settings of the hyperparameters introduced in our method.
The loss weights $\lambda_{\rm sem}$, $\lambda_{\rm bce}$, $\lambda_{\rm dice}$, $\lambda_{\rm cls}$, and $\lambda_{\rm text}$ in the overall training objective follow the default configuration of the original PIXAR protocol~\cite{shang2026masks}.
For balanced mini-batch sampling, we set the real-image ratio to $\rho=0.5$, so that each mini-batch contains an equal number of real and tampered images.
For computational efficiency, ablations are conducted on a reduced training set of $7{,}136$ Qwen-Image and $400$ Gemini-2.5 tampered images with all other dataset and training settings inherited unchanged.

\subsection{Details of PIXAR \cite{shang2026masks}}
\label{app_sec:pixar}
PIXAR is the first benchmark designed for precise pixel-level tampering detection on locally edited images.
Specifically, the PIXAR training set contains 380K samples covering eight editing operations, all generated by Qwen-Image VLMs~\cite{wu2025qwen}, whereas its test set covers the same eight operations produced by six state-of-the-art generative models, including Qwen-Image VLMs~\cite{wu2025qwen}, GPT-Image-1.5~\cite{gptimage15}, Gemini-3~\cite{gemini3}, FLUX.2~\cite{flux2}, Seedream~4.5~\cite{seedream2025seedream40}, and Gemini-2.5~\cite{gemini25}. 
This multi-generator test set makes PIXAR a natural benchmark for studying cross-generator generalization.
To further assess generalization to the latest generative models, we augment test set with samples from GPT-Image-2.0~\citep{openai2026gptimage2} and Gemini-3.1~\citep{googledeepmind2026gemini31procard}.

Each sample in PIXAR is provided with rich metadata, consisting of (i) a real source image, (ii) its tampered counterpart, (iii) the recommended binary pixel-level label map $\mathbf{M}_{\tau}$ from a default $\tau$, and (iv) the raw per-pixel difference map from which alternative labels for other $\tau$ values can be derived.
Following the protocol of PIXAR~\cite{shang2026masks}, we set $\tau=0.05$ to obtain pixel-level labels for both the training and test datasets.

\subsection{Test Dataset Distribution}
\label{app_sec:test_distribution}
\begin{figure}
    \centering
    \includegraphics[width=1\linewidth]{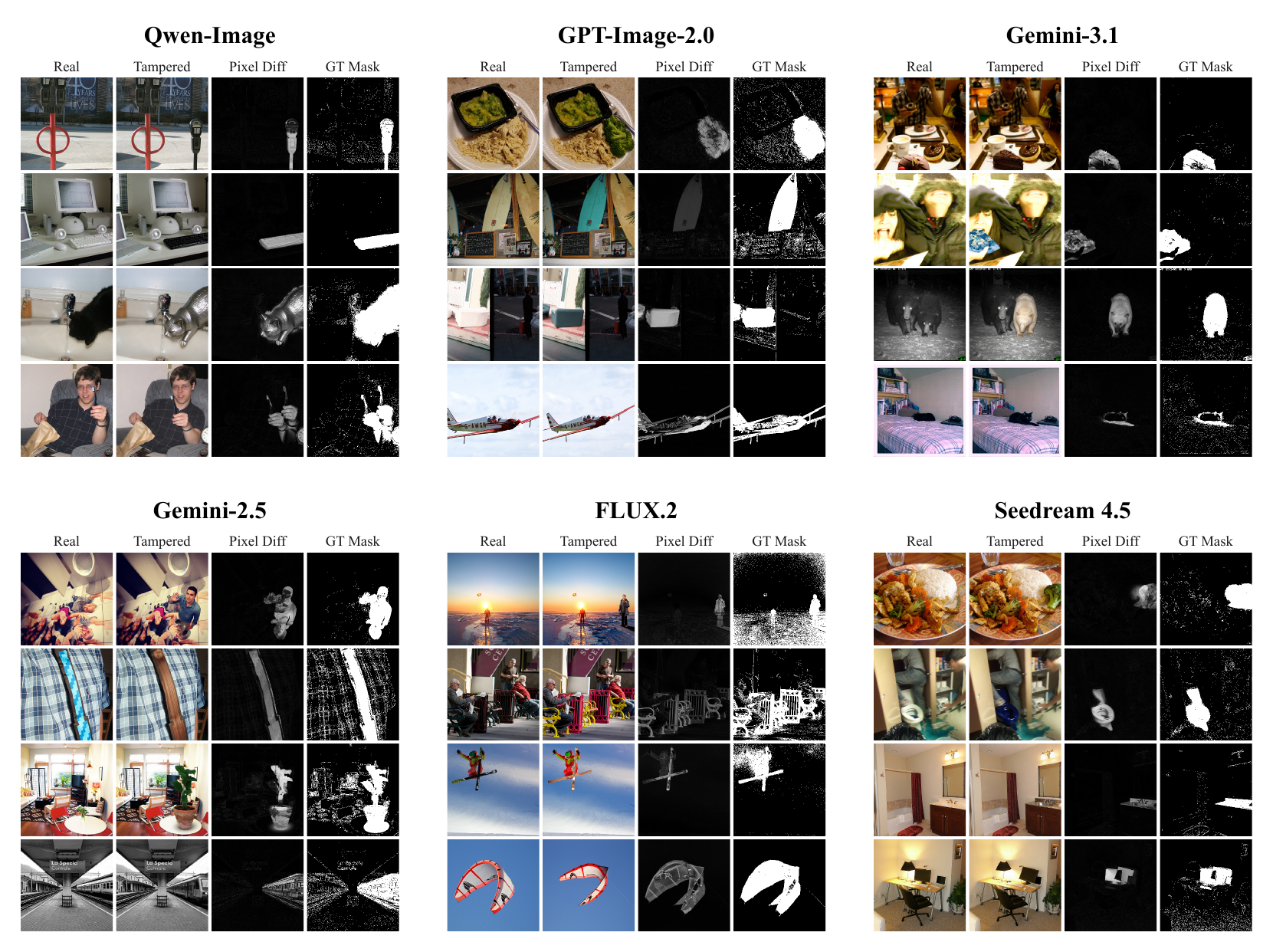}
    \caption{\textbf{Visualizations of tampered images used in test set}.}
    \label{fig:model_grid}
\end{figure}

\begin{wraptable}{r}{0.35\columnwidth}
    \caption{Test-dataset distribution under the new composition.}
    \label{tab:app_test_distribution}
    \centering
    \resizebox{1\linewidth}{!}{
    \begin{tabular}{lc}
        \toprule
        Generator & \# tampered test samples \\
        \midrule
        Qwen-Image       & 742 \\
        Gemini-2.5       & 468 \\
        \midrule
        GPT-Image-2.0    & 800 \\
        Gemini-3.1       & 783 \\
        FLUX.2           & 671 \\
        Seedream~4.5     & 513 \\
        \midrule
        Total            & 3977 \\
        \bottomrule
    \end{tabular}}
\end{wraptable}
We show the test dataset distribution in \tabref{tab:app_test_distribution}.
Under our evaluation protocol, Qwen-Image~\cite{wu2025qwen} and Gemini-2.5~\cite{gemini25} act as the in-domain (ID) generators, while the remaining four models, GPT-Image-2.0~\cite{openai2026gptimage2}, Gemini-3.1~\cite{googledeepmind2026gemini31procard}, FLUX.2~\cite{flux2}, and Seedream~4.5~\cite{seedream2025seedream40}, are held out and treated as out-of-distribution (OOD) generators unseen during training.
\figref{fig:model_grid} illustrates the composition of our tampering benchmark across six modern image generators. 

Following PIXAR~\citep{shang2026masks}, the Qwen-Image~\cite{wu2025qwen}, Gemini-2.5~\cite{gemini25}, FLUX.2~\cite{flux2}, and Seedream~4.5~\cite{seedream2025seedream40} subsets reuse the same set of generators.

To further assess the ability of \ours to keep pace with the most recent image generators, we additionally include two latest-generation models, GPT-Image-2.0~\cite{openai2026gptimage2} and Gemini-3.1~\cite{googledeepmind2026gemini31procard}. 

\subsection{In-Domain Evaluation}
\label{app_sec:in_domain}
We report in-domain experimental results in~\tabref{tab:main_results_id}, covering both pixel-level localization (gIoU, cIoU) and binary classification accuracy. 
Since LISA is primarily designed as a segmentation model and does not include a dedicated binary real-vs-tampered classification head, we do not report its binary classification accuracy.
At the 13B scale, \ours-13B substantially improves in-domain binary classification accuracy (84.9\% vs.\ 69.1\% for PIXAR-13B), while remaining comparable to PIXAR-13B on pixel-level gIoU and cIoU.

\begin{table}[ht]
    \caption{\textbf{Pixel-level localization and binary classification results on the two in-domain (ID) generators.} ``Avg.'' denotes the average results across the two generators.}
    \label{tab:main_results_id}
    \centering
    \renewcommand{\arraystretch}{1.1}
    \resizebox{.9\linewidth}{!}{
    \begin{tabular}{l|ccc|ccc|ccc}
        \toprule
        \multicolumn{1}{c|}{\multirow{2}{*}{Method}} &
        \multicolumn{3}{c|}{Per-generator gIoU} &
        \multicolumn{3}{c|}{Per-generator cIoU} &
        \multicolumn{3}{c}{Per-generator Binary Accuracy (\%)} \\
        \cmidrule(lr){2-4}\cmidrule(lr){5-7}\cmidrule(lr){8-10}
        & Qwen-Image & Gemini-2.5 & Avg. & Qwen-Image & Gemini-2.5 & Avg. & Qwen-Image & Gemini-2.5 & Avg. \\
        \midrule
        SIDA-7B    & 0.146 & 0.128 & 0.137 & 0.152 & 0.131 & 0.141 & 22.4  & 18.8  & 20.6 \\
        PIXAR-7B   & 0.222 & 0.154 & 0.188 & 0.263 & 0.160 & 0.211 & 94.7  & 57.3  & 76.0 \\
        \rowcolor[HTML]{F2F2F2}
        \ours-7B   & 0.204 & 0.229 & 0.217 & 0.227 & 0.248 & 0.237 & 96.2  & 81.2  & 88.7 \\
        \midrule
        SIDA-13B   & 0.112 & 0.084 & 0.098 & 0.128 & 0.091 & 0.109 & 41.0  & 30.1  & 35.5 \\
        PIXAR-13B  & 0.219 & 0.213 & 0.216 & 0.241 & 0.226 & 0.233 & 93.3  & 44.9  & 69.1 \\
        \rowcolor[HTML]{F2F2F2}
        \ours-13B  & 0.189 & 0.238 & 0.213 & 0.207 & 0.254 & 0.231 & 95.3  & 74.6  & 84.9 \\
        \bottomrule
    \end{tabular}
    }
\end{table}

\begin{table}[!b]
    \caption{\textbf{Source composition study: Full and three-source compositions.}
    Each row represents a different three-source training composition built on top of Qwen-Image.
    \textit{\#Src.} denotes the number of training sources.
    Per-generator columns report tampered accuracy (\%) on each test generator.
    \textit{Avg. (All)} denotes the average accuracy across all six generators;
    \textit{Avg. (OOD)} denotes the average accuracy across the held-out OOD generators.
    \textbf{Bold} indicates the best result, and \underline{underline} indicates the second-best result,
    both within the three-source composition family.}
    \label{tab:source_composition_three}
    \centering
    \renewcommand{\arraystretch}{1.1}
    \resizebox{\linewidth}{!}{
    \begin{tabular}{l|c|cccccc|cc}
        \toprule
        \multirow{2}{*}{Training composition} & \multirow{2}{*}{\#Src.} &
        \multicolumn{6}{c|}{Per-generator Accuracy (\%)} &
        \multicolumn{2}{c}{Average} \\
        \cmidrule(lr){3-8}\cmidrule(lr){9-10}
        & & Qwen-Image & GPT-Image-2.0 & Gemini-3.1 & Gemini-2.5 & FLUX.2 & Seedream 4.5 & All & OOD \\
        \midrule
         \multicolumn{10}{l}{\textit{Full mixture}} \\
        All Generators
            & 6 & 84.8 & 88.5 & 91.2 & 87.6 & 83.2 & 91.6 & 87.8 & --- \\
        \midrule\midrule
        \multicolumn{10}{l}{\textit{Three-source compositions}} \\
        Qwen-Image + Gemini-2.5 + FLUX.2
            & 3 & 78.0 & 74.0 & 73.8 & \underline{77.6} & 80.0 & 70.4 & 75.6 & 72.7 \\
        Qwen-Image + Gemini-2.5 + GPT-Image-2.0
            & 3 & \textbf{82.1} & \textbf{87.4} & \underline{80.0} & \textbf{83.1} & 62.7 & \underline{80.7} & \textbf{79.3} & \textbf{74.5} \\
        Qwen-Image + Gemini-3.1 + FLUX.2
            & 3 & \underline{79.8} & 70.7 & \textbf{84.0} & 76.1 & 80.8 & 75.6 & \underline{77.8} & \underline{74.1} \\
        Qwen-Image + GPT-Image-2.0 + FLUX.2
            & 3 & 78.0 & \underline{82.8} & 27.4 & 34.4 & \textbf{82.1} & 71.7 & 62.7 & 44.5 \\
        Qwen-Image + Seedream 4.5 + FLUX.2
            & 3 & 79.0 & 73.9 & 36.9 & 37.6 & \underline{82.0} & \textbf{81.5} & 65.1 & 49.5 \\
        \bottomrule
    \end{tabular}
    }
\end{table}

\subsection{More Sources Composition Experiments}
\label{app_sec:more_source_composition}
We further report results for three-source compositions and the full-mixture setting in \tabref{tab:source_composition_three}.
Three-source compositions are expected to achieve stronger generalization than two-source compositions, as they expose the model to a broader range of generator-specific artifacts during training. However, they also require access to additional training domains and therefore represent a less resource-efficient setting than our two-source design.
Since the full-mixture configuration is trained on samples from all six generators, no generator remains held out for evaluating Avg.~(OOD). We therefore report its overall performance as a reference point, which can be interpreted as an approximate upper bound within the available generator pool.

\subsection{Additional Qualitative Results}
\label{app_sec:additional_qualitative}
\begin{figure}[!h]
    \centering
    \includegraphics[width=1\linewidth]{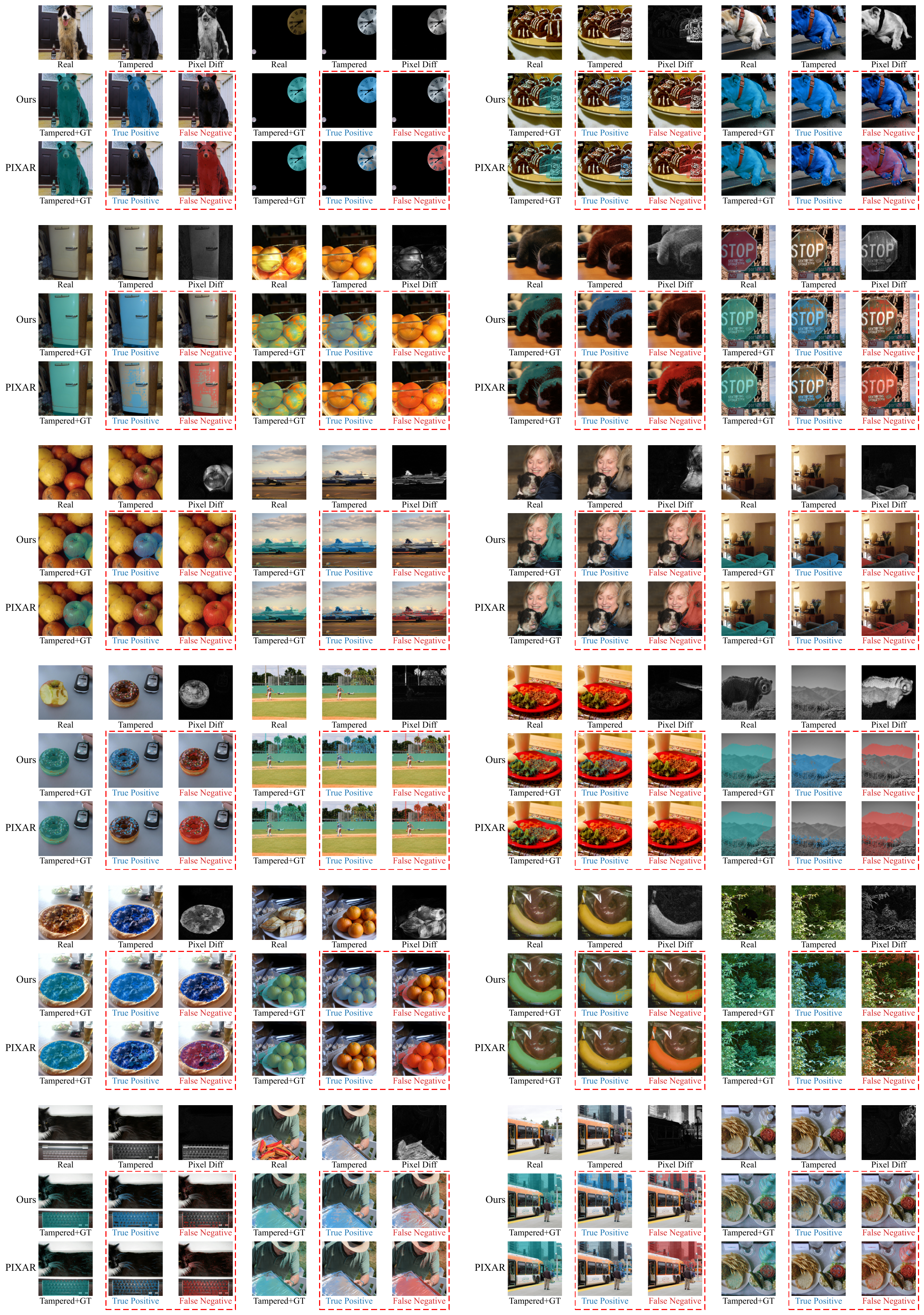}
    \caption{\textbf{More qualitative comparisons of predicted tampered pixels between \ours and PIXAR~\citep{shang2026masks}}.}
    \label{fig:app_more_visualization}
    \vspace{-15pt}
\end{figure}

We provide additional qualitative results in~\figref{fig:app_more_visualization} to further compare \ours with PIXAR~\citep{shang2026masks}.
The examples cover GPT-Image-1.5, Gemini-3, FLUX.2, and Seedream~4.5, and include diverse tampering types and object categories.
Across these cases, \ours more accurately localizes the edited regions and produces fewer false-positive responses on visually plausible but unedited areas, demonstrating stronger cross-generator robustness.

\subsection{Different Ratios between Real and Tampered}
\label{app_sec:different_ratio}
We vary the per-batch real:tampered ratio enforced by MiniBatch Sampling, keeping all other training settings fixed; results are reported in~\tabref{tab:ablation_bt_ratio}.
Among the three configurations, the balanced 1:1 ratio delivers the strongest overall performance, achieving the highest average accuracy (66.3\%) and pixel-level F1 score (29.89), while remaining competitive in per-generator accuracy across all four out-of-distribution generators.

\begin{table}[ht]
    \caption{\textbf{Ablation on the balanced-training real:tampered ratio.} All variants share identical training data, model, and hyperparameters.}
    \label{tab:ablation_bt_ratio}
    \centering
    \renewcommand{\arraystretch}{1.1}
    \resizebox{0.62\linewidth}{!}{
        \begin{tabular}{c|cccc|cc}
        \toprule
        \multirow{2}{*}{Real:Tampered} & \multicolumn{4}{c|}{Per-generator Accuracy (\%)} & \multicolumn{2}{c}{Average} \\
        \cmidrule(lr){2-5}\cmidrule(lr){6-7}
        & GPT-Image-2.0 & Gemini-3.1 & FLUX.2 & Seedream 4.5 & Acc. & Pixel F1 \\
        \midrule
        2:1 & 58.6 & 39.8 & 42.9 & 72.7 & 53.5 & 23.68 \\
        \rowcolor[HTML]{F2F2F2}
        1:1 & \textbf{71.5} & \textbf{56.6} & \textbf{53.2} & \textbf{83.8} & \textbf{66.3} & \textbf{29.89} \\
        1:2 & 52.7 & 33.5 & 38.5 & 71.3 & 49.0 & 26.37 \\
        \bottomrule
        \end{tabular}
    }
\end{table}

\end{document}